%% file: main.tex
\documentclass[10pt,twocolumn,letterpaper]{article}
\input{public_preamble}
\input{authors}
\title{\CityCodeTitle}
\author{\CityCodeArxivAuthorBlock}
\hypersetup{pdftitle={\CityCodeTitle},pdfauthor={\CityCodeArxivAuthorMetadata}}
\begin{document}
\input{sections/main_body}

\input{sections/llm_usage}
{\small\bibliography{bibliography/references}}
\bibliographystyle{ieeenat_fullname}
\end{document}

%% file: public_preamble.tex
\usepackage[T1]{fontenc}
\usepackage[pagenumbers]{cvpr}
\input{math_commands.tex}

\usepackage{amsmath,amssymb}
\usepackage{booktabs}
\usepackage{graphicx}
\usepackage{array,rotating}
\newcommand{\rev}[1]{#1}
\newenvironment{revblock}{}{}
\usepackage{placeins}
\definecolor{cvprblue}{rgb}{0.21,0.49,0.74}
\usepackage[pagebackref,breaklinks,colorlinks,allcolors=cvprblue]{hyperref}
\usepackage{url}

\newcommand{\CityCodeTitle}{CityCode: Building Agent-Operable 3D Cities from Sparse Aerial Observations}
\newcommand{\CityCodeVersionNote}{%
  \par\smallskip
  {\normalfont\small\itshape Main paper only. References to supplementary material follow its original numbering.\par}%
}

%% file: math_commands.tex
\usepackage{amsmath,amsfonts,bm}

\def\eqref#1{equation~\ref{#1}}

\def\1{\bm{1}}

\DeclareMathAlphabet{\mathsfit}{\encodingdefault}{\sfdefault}{m}{sl}
\SetMathAlphabet{\mathsfit}{bold}{\encodingdefault}{\sfdefault}{bx}{n}

%% file: authors.tex
\newcommand{\CityCodeArxivAuthorBlock}{%
  \parbox{0.94\textwidth}{\centering
    \textbf{Jiahui Sun\textsuperscript{1,2,*},
      Ran Bi\textsuperscript{1,*},
      Yifei Liu\textsuperscript{2,6,*},
      Xiaolong Zhou\textsuperscript{2,5}}\\[0.35em]
    \textbf{Yishuo Xu\textsuperscript{1},
      Yuanqi Li\textsuperscript{1},
      Congcong Zhu\textsuperscript{3},
      Lei Yang\textsuperscript{4}}\\[0.35em]
    \textbf{Yanwen Guo\textsuperscript{1,\textdagger},
      Zhihang Zhong\textsuperscript{5,\textdagger},
      Hongjie Zhang\textsuperscript{2,\textdagger}}\\[0.65em]
    {\normalfont\small
      \textsuperscript{1}Nanjing University\quad
      \textsuperscript{2}Shanghai Artificial Intelligence Laboratory\\
      \textsuperscript{3}University of Science and Technology of China\quad
      \textsuperscript{4}SenseTime\\
      \textsuperscript{5}Shanghai Jiao Tong University\quad
      \textsuperscript{6}Beihang University\\[0.55em]
      \textsuperscript{*}Equal contribution.\quad
      \textsuperscript{\textdagger}Co-corresponding authors.}\CityCodeVersionNote%
  }%
}
\newcommand{\CityCodeArxivAuthorMetadata}{Jiahui Sun, Ran Bi, Yifei Liu, Xiaolong Zhou, Yishuo Xu, Yuanqi Li, Congcong Zhu, Lei Yang, Yanwen Guo, Zhihang Zhong, Hongjie Zhang}

%% file: sections/main_body.tex
\twocolumn[{%
  \renewcommand\twocolumn[1][]{#1}%
  \maketitle
  \vspace{-1.8em}
  \begin{center}
    \includegraphics[width=\textwidth]{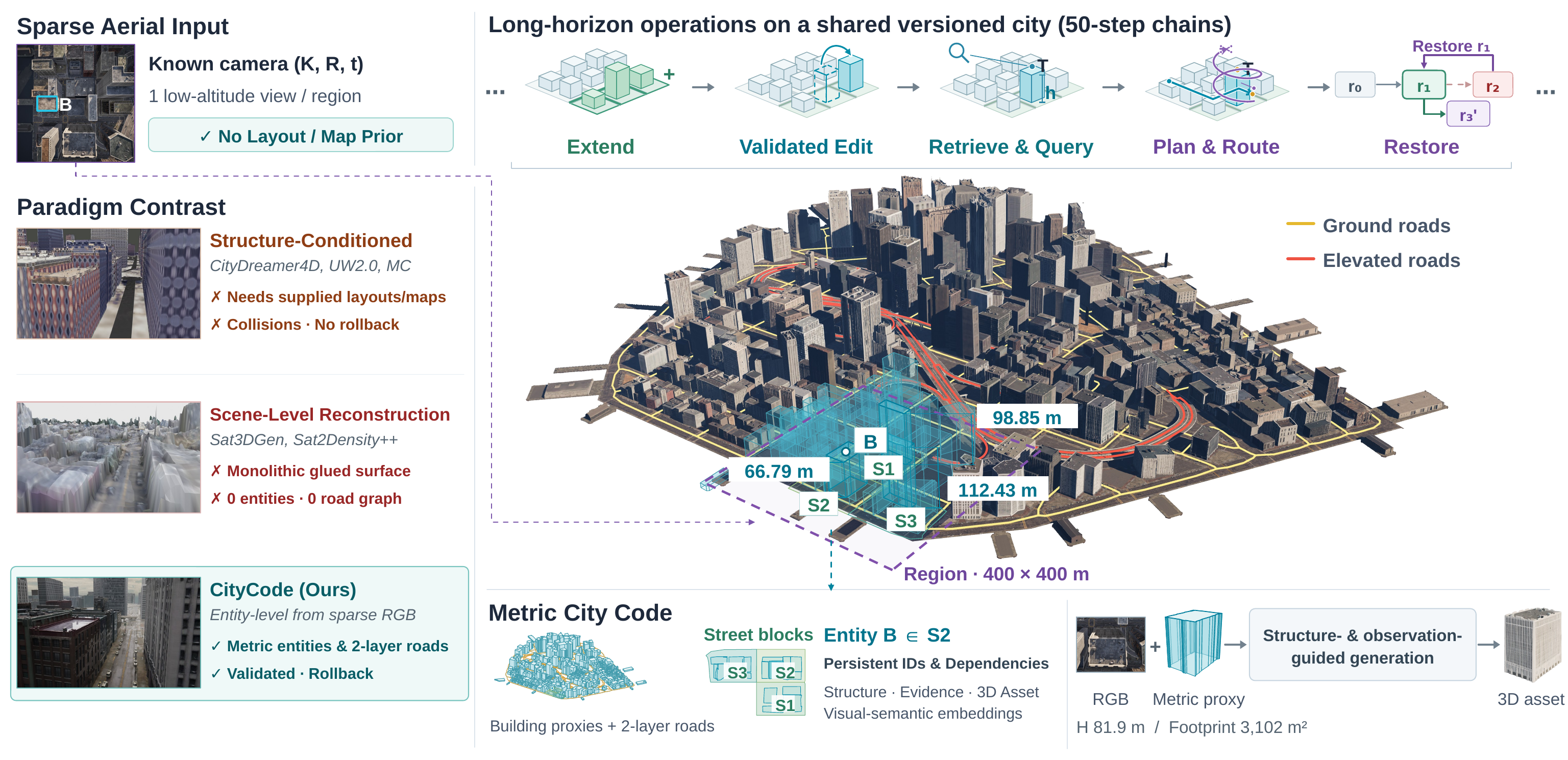}
    \vspace{-0.55em}
    \captionof{figure}{\textbf{From sparse aerial observations to agent-operable 3D cities.} Unlike structure-conditioned generators requiring external layouts or maps and image-to-scene methods producing uneditable monolithic surfaces, \textbf{Air2Code} recovers metric building instances and layered road topology from sparse aerial RGB views to drive structure- and observation-guided 3D generation. \textbf{CityCode} maintains these entities in a shared, versioned state for long-horizon querying, validated editing, routing, and restoration (see supplementary video).}
    \label{fig:teaser}
    \vspace{0.15em}
  \end{center}%
}]

\begin{abstract}
  \input{sections/00_abstract}

\end{abstract}

\input{sections/01_introduction}
\input{sections/02_related_work}
\input{sections/03_method}
\input{sections/04_experiments}

\FloatBarrier
\input{sections/05_conclusion}

%% file: sections/00_abstract.tex
Urban spatial intelligence calls for 3D cities that agents can measure, query, and modify across successive tasks, yet structure-conditioned generators rely on external layouts while image-to-scene methods produce uneditable monolithic surfaces. We formulate grounded urban world construction: generating instance-level parameterized 3D cities from one low-altitude aerial RGB image per region and camera parameters without external maps. We introduce CityCode, an agent-operable representation linking observation evidence, metric building proxies, layered road-block topology, semantic embeddings, and 3D assets through persistent entity identities, with dependency-aware versioned execution for validated edits and instant state restoration. From sparse views, Air2Code pairs visual structure parsing with CityCode-Depth, our adapted aerial depth model, to recover metric building instances and road graphs, resolving single-view occlusion through structure- and observation-guided generation via structure-guided appearance completion, CityCode-Building (our structure-aware 3D building generator), and height-guided background synthesis. Across CS-Aerial and UrbanScene3D, CityCode leads on 21 of 22 metric-geometry and visual-quality metrics while preserving geometric fidelity with open-weight backbones. In long-horizon evaluation, CityCode completes all 2,500 steps across 50 task chains and outperforms agent-scripted baselines with zero constraint violations at 3.0--3.6$\times$ lower agent time, turning sparse aerial observations into persistent, agent-operable urban worlds. Code will be publicly available.

%% file: sections/01_introduction.tex
\section{Introduction}
\label{sec:introduction}

Urban spatial intelligence requires agents that can measure, query, and modify a target 3D city across successive tasks---locating buildings, reasoning over road access, and updating layouts while keeping metric geometry, road connectivity, and visual semantics synchronized~\citep{zhang2026generativeurban,caldarelli2023complexity}. As edits accumulate, later queries and plans must stay consistent with the evolving spatial state, motivating \emph{agent-operable 3D cities} constructed from real-scene observations and maintained over long horizons.

Constructing such cities from lightweight input remains challenging. Structure-conditioned generators either synthesize layouts from text or design rules (CityGenAgent, CityCraft, MajutsuCity~\citep{liu2026citygenagent,deng2024citycraft,huang2025majutsucity}) or rely on external maps and height layouts (UrbanWorld, GaussianCity, CityDreamer4D, Map2World~\citep{shang2024urbanworld,wang2025urbanworld2,wang2025sat2realcity,xie2025gaussiancity,xie2026citydreamer4d,chung2026map2world}). Synthesized layouts are not grounded in a target city, while public maps often suffer from missing heights, incomplete coverage, or outdated geometry~\citep{herfort2023osmcompleteness,bernard2022buildingheight,juhasz2021temporalaccuracy}---and neither maintains a versioned, constraint-validated state across edits. Sparse aerial RGB imagery instead offers a timely, accessible observation source from which target-city structure can be recovered~\citep{remondino2011uav}.

Conversely, image-based scene reconstruction and generation methods (Sat2City v2, Sat2Density++, Sat3DGen, MagicCity, SceneFuse-3D~\citep{hua2026sat2cityv2,qian2026sat2densitypp,qian2026sat3dgen,yao2025magiccity,zheng2025scenefuse3d}) output monolithic surfaces or radiance fields, while monocular building reconstructors recover only untextured LoD1 boxes without road graphs or operable state~\citep{li2024mlsbrn,shao2024singlerecon,hu2026dgbrf}. Moreover, sparse aerial views lack direct metric scale and leave most facades occluded; simply chaining off-the-shelf vision and 3D models distorts building proportions and scene alignment unless generative completion is tightly coupled with recovered metric structure.

We therefore introduce \emph{grounded urban world construction}: given one low-altitude aerial RGB image per region and camera parameters, we construct an instance-level parameterized 3D city for continued spatial reasoning and action (Figure~\ref{fig:teaser}). Our representation, \textbf{CityCode}, uses persistent entity identities to link visual evidence, metric LoD1 proxies, road-block topology, semantic embeddings, and 3D assets across revisions, while dependency-aware versioned execution validates edits, updates dependent records, and supports queries, routing, and state restoration.

To construct CityCode from sparse views, \textbf{Air2Code} couples metric structure recovery with \emph{structure- and observation-guided generation} (Figure~\ref{fig:method-overview}). Visual parsing and CityCode-Depth first recover metrically scaled building proxies and layered road graphs; these proxies then anchor unobserved facade synthesis via structure-guided appearance completion and CityCode-Building, our structure-aware 3D building generator, alongside height-guided background generation, binding each asset only after placement checks pass. Our contributions are as follows:

\begin{revblock}
\begin{itemize}
  \item \textbf{Setting and representation.} We introduce grounded urban world construction from sparse aerial observations, and CityCode, to our knowledge the first city-scale parametric representation with recovered metric building instances and urban structure. These entities are kept as a versioned state that agents can operate on.
  \item \textbf{Reusable models and a structure- and observation-guided generation mechanism.} We develop two reusable components: CityCode-Depth, an aerial depth model adapted for metric recovery, and CityCode-Building, a structure-aware 3D building generator adapted to preserve target footprints and heights. Air2Code couples these components through structure- and observation-guided generation: recovered metric entities constrain geometry, while image evidence guides appearance completion and 3D synthesis.
  \item \textbf{Evaluation and long-horizon use.} CityCode leads on 21 of 22 metric-geometry and visual-quality metrics across CS-Aerial and UrbanScene3D, maintains consistent state across 50 chains of 50 steps, and outperforms agent-scripted baselines with zero constraint violations at 3.0--3.6$\times$ lower agent time.
\end{itemize}
\end{revblock}

%% file: sections/02_related_work.tex
\section{Related Work}
\label{sec:related-work}

\subsection{Urban Modeling and Generation}
\label{sec:related-city-generation}

\textbf{Design-driven and structure-conditioned generation.}
Prior city generators synthesize layouts and assets from design specifications~\citep{liu2026citygenagent,deng2024citycraft,huang2025majutsucity}, external 2D/OSM maps and street-view cues~\citep{zhang2024cityx,shang2024urbanworld,wang2025urbanworld2,chung2026map2world}, or supplied semantic and height maps~\citep{xie2024citydreamer,xie2025gaussiancity,xie2026citydreamer4d,wang2025sat2realcity,hua2025sat2city}. While enabling layout-level control, all assume urban structure is synthesized or externally provided rather than recovered metrically from target-city observations, and none maintain a versioned state across edits.

\textbf{Scene-level and building-only reconstruction from imagery.}
Satellite/aerial methods reconstruct monolithic, uninstantiated scene surfaces, radiance fields, or Gaussian splats via view synthesis and 3D inpainting~\citep{hua2026sat2cityv2,qian2026sat2densitypp,qian2026sat3dgen,yao2025magiccity,zheng2025scenefuse3d}, whereas monocular building reconstructors recover only untextured LoD1 boxes from off-nadir imagery~\citep{li2024mlsbrn,shao2024singlerecon,hu2026dgbrf}. Neither line jointly recovers building instances, layered road connectivity, and entity-linked 3D assets in operable cities (Table~S22, Appendix~J).

\subsection{Structured Representations and Spatial Interaction}
\label{sec:related-spatial-intelligence}

\begin{figure*}[t]
  \centering
  \includegraphics[width=\linewidth]{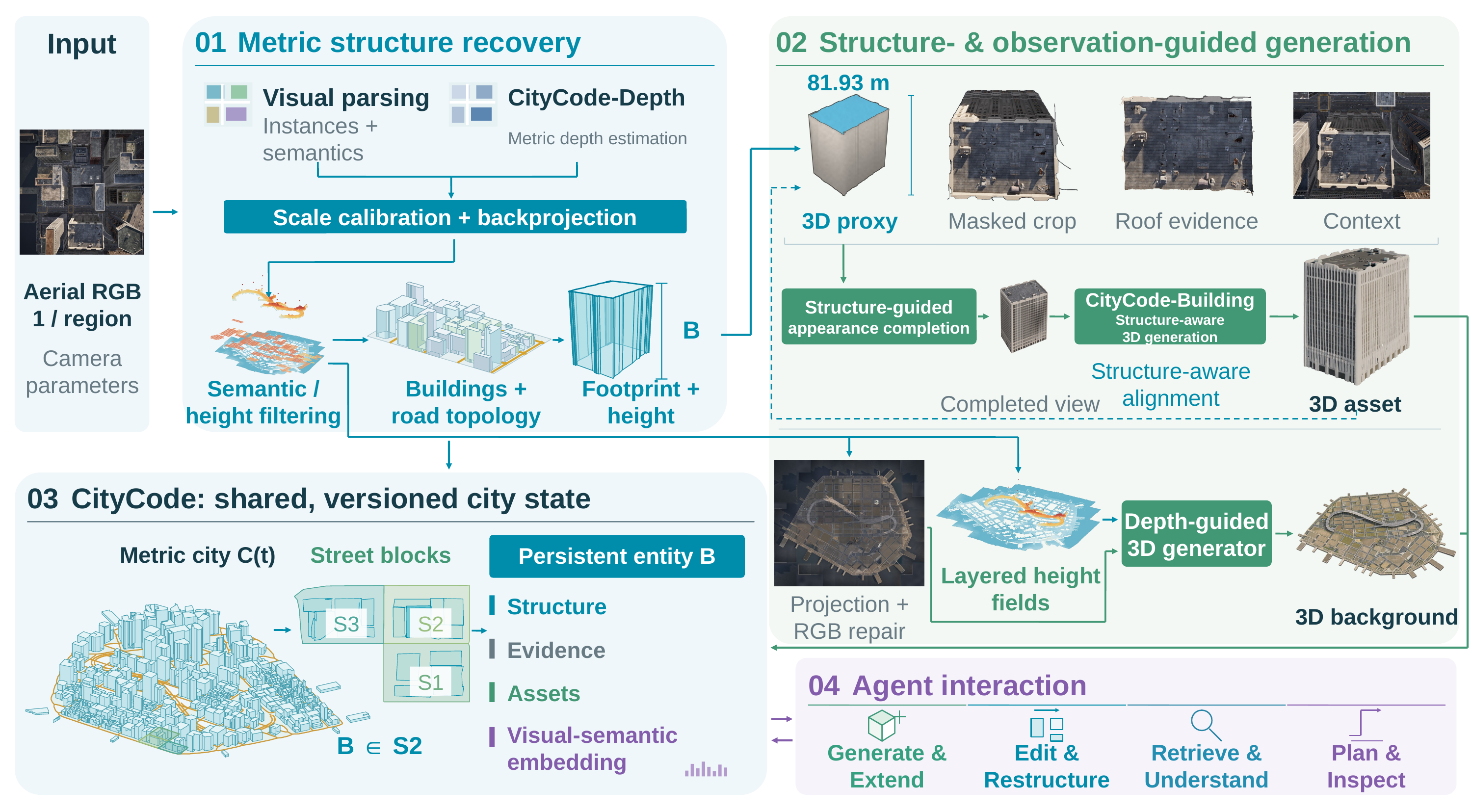}
  \caption{\textbf{Overview of CityCode and Air2Code.} Air2Code uses CityCode-Depth for metric depth estimation and CityCode-Building for structure-aware 3D building generation, coupling recovered geometry and visual evidence through structure- and observation-guided generation. CityCode links structure, evidence, assets, and visual-semantic embeddings through persistent entity identities in a shared, versioned city state.}
  \label{fig:method-overview}
\end{figure*}

Semantic city standards such as CityGML~3.0 define object schemas and versioning for static GIS exchange~\citep{ogc2021citygml}, while 3D scene graphs and embodied agents support perception, grounding, and planning in indoor or robotic settings~\citep{mao2025spatiallm,hughes2022hydra,gu2023conceptgraphs,rana2023sayplan,huang2024leo}. At the urban scale, recent benchmarks study spatial reasoning, decision-making, and navigation over pre-built environments~\citep{feng2025citybench,xu2026citycube,wu2025metaurban}. Meanwhile, code-backed world models decouple symbolic state updates or scene programs from visual asset generation mainly for indoor or object-centric scenes~\citep{chen2026codeworldmodel,wang2026scenecode,li2026recursive}. CityCode bridges observation-grounded urban reconstruction and code-maintained world state, supporting revision-bound queries, clearance-checked dependent edits, and historical restoration across long task horizons.

%% file: sections/03_method.tex
\section{Method}
\label{sec:method}

\subsection{CityCode Representation}
\label{sec:citycode-representation}

\paragraph{Metric urban structure.}
CityCode organizes buildings, roads, blocks, and environment layers in a shared metric frame (Figure~\ref{fig:teaser}; notation in Appendix~B.1). Building footprints, base elevations, and heights define LoD1 proxies. Roads form a spatial graph with centerlines, widths, and connectivity; blocks and spatial relations encode building membership, neighborhoods, and road associations. Environment layers store registered background imagery with height and confidence fields, and OSM XML/CityJSON projections support semantic 3D interchange (Appendix~H).

\paragraph{Entity representation.}
For building $i$ at revision $r$, the linked records are summarized as
\begin{equation}
    b_i^{(r)} = \bigl(\mathrm{id}_i,\; P_i^{(r)},\; z_i^{(r)},\; h_i^{(r)},\;
    \mathcal{O}_i,\; a_i^{(r)},\; T_i^{(r)},\; \mathcal{E}_i^{(r)}\bigr),
    \label{eq:citycode-building}
\end{equation}
where $P_i$ is the building footprint polygon and $(z_i,h_i)$ are its base elevation and height; $\mathcal{O}_i$ references observations and construction provenance; and $(a_i,T_i)$ bind a textured asset and its local-to-world transform. Generated completions are recorded separately from observations, and entity identity persists when assets or placements change. The structural footprint $P_i$ remains distinct from the transformed mesh bounds used for placement checks, while visual-semantic embeddings $\mathcal{E}_i^{(r)}$ from rendered asset views support text and image retrieval.

\paragraph{Versioned city state.}
An immutable city revision stores entities, the road graph, blocks, spatial relations, environment layers, and asset bindings. Entity lookup and derived appearance indexes are revision-bound. Unchanged evidence and assets are shared across revisions. Section~\ref{sec:versioned-execution} describes updates, validation, and restoration.

\subsection{Air2Code: Construction from Sparse Aerial Observations}
\label{sec:air2code}

Given one aerial RGB image $I$ per region with intrinsics $K$ and pose $(R,t)$, Air2Code constructs CityCode in two coupled stages (Figure~\ref{fig:method-overview}): recovering metric urban entities and using them alongside observations to guide 3D generation.

\subsubsection{Metric Urban Structure Recovery}
\label{sec:metric-structure-recovery}

\paragraph{CityCode-Depth: depth adaptation and scale calibration.}
A visual parser extracts building-instance and background-semantic masks (Gemini~3 Pro Image by default, or open-weight SAM~3.1; Section~\ref{sec:ablations-analysis}), with road and hardscape pixels defining ground anchors $\Omega$. For \emph{depth adaptation}, we fine-tune MoGe2-Aerial~\citep{song2026aerialmetric} on RGB--depth pairs from a disjoint synthetic map using an affine-invariant point-map loss and metric-scale supervision, to obtain CityCode-Depth, which predicts metrically scaled camera-forward depth $\hat d$. When local ground elevation is available, camera height $H_{\mathrm{cam}}(u)$ above $u\in\Omega$ and ray $q(u)=K^{-1}\tilde u$ provide lightweight \emph{scale calibration} before backprojection (Appendix~C.4):
\begin{equation}
    \begin{aligned}
    d_{\mathrm{ref}}(u)&=-\frac{H_{\mathrm{cam}}(u)}{(Rq(u))_z},
    &s&=\operatorname*{median}_{u\in\Omega}\frac{d_{\mathrm{ref}}(u)}{\hat d(u)},\\
    X(u)&=R\bigl(s\hat d(u)q(u)\bigr)+t.
    \end{aligned}
    \label{eq:metric-backprojection}
\end{equation}
\paragraph{Instance-aware building recovery.}
Backprojected points retain their instance and semantic labels. In a metric BEV grid, stable height bands with connected support guide footprint cleanup and roof-elevation estimation. Polygonizing the cleaned support yields $P_i$; the highest accepted roof elevation above base $z_i$ gives height $h_i$, defining a LoD1 proxy. Overlapping regional observations are associated with global building identities while retaining source links (Appendix~C.6).

\paragraph{Layered background and road structure.}
Background observations are fused into ground and elevated height fields, allowing ground and bridge decks to coexist at the same XY location, with validity and confidence retained for generation. Road masks are skeletonized and vectorized into junctions and centerlines; the network defines blocks and building--road relations.

\subsubsection{Structure- and Observation-Guided Generation}
\label{sec:building-generation}

Sparse aerial views leave most facades and under-deck areas occluded. Each entity's recovered proxy $(P_i,h_i)$ and aerial crops therefore jointly guide \emph{structure-guided appearance completion} and \emph{CityCode-Building}, while layered height fields guide background generation.

\paragraph{Structure-guided appearance completion.}
Directly lifting truncated aerial crops into 3D distorts building geometry. We formulate \emph{structure-guided appearance completion} $C$ (GPT-image-2 by default, or open-weight FLUX.2-klein-9B) to synthesize an isolated, complete oblique view $J_i$:
\begin{equation}
J_i=C\bigl(\operatorname{Render}(P_i,h_i),I_i^{\mathrm{src}},I_i^{\mathrm{roof}},I_i^{\mathrm{ctx}},h_i\bigr).
\label{eq:building-generation}
\end{equation}
Here \emph{structure guidance}---rendered LoD1 proxy $\operatorname{Render}(P_i,h_i)$ and height $h_i$---anchors 3D massing and proportions, while \emph{observation evidence}---masked source crop $I_i^{\mathrm{src}}$, BEV roof crop $I_i^{\mathrm{roof}}$, and surrounding context $I_i^{\mathrm{ctx}}$---grounds facade and roof appearance.

\paragraph{CityCode-Building: structure-aware 3D generation.}
Even from a completed 2D view $J_i$, an off-the-shelf 3D generator suffers from single-view vertical scale ambiguity. We develop \emph{CityCode-Building}, our structure-aware 3D building generator, by adapting the sparse-structure, geometry, and material stages of TRELLIS.2~\citep{xiang2025trellis2} on curated SAT-SKYLINES building assets~\citep{jin2025satskylines}. During Stage-1 sparse-structure training, a frozen decoder maps predicted clean latents to occupancy probabilities whose vertical-column maxima define a soft footprint $\hat F$, supervised against canonical footprint $F$ and aspect ratio $H/\sqrt{A}$:
\begin{equation}
\begin{aligned}
\mathcal L_{\mathrm{structure}}={}&\mathcal L_{\mathrm{FM}}+\alpha(k)\bigl[
\lambda_{\mathrm{bce}}\mathcal L_{\mathrm{balBCE}}(\hat F,F)\\
&+\lambda_{\mathrm{iou}}\mathcal L_{\mathrm{IoU}}(\hat F,F)
+\lambda_{\mathrm{ratio}}\mathcal L_{\mathrm{ratio}}\bigr],
\end{aligned}
\label{eq:footprint-supervision}
\end{equation}
where $\mathcal L_{\mathrm{balBCE}}$ and $\mathcal L_{\mathrm{IoU}}$ supervise horizontal support, $\mathcal L_{\mathrm{ratio}}$ penalizes log-ratio error in $H/\sqrt{A}$, and $\alpha(k)$ warms up these \emph{auxiliary structure losses} over step $k$ (Appendix~D.4); Stages~2--3 retain flow matching $\mathcal L_{\mathrm{FM}}$.

At inference (Figure~\ref{fig:method-overview}), $J_i$ conditions Stage-1 sparse-structure sampling, which already captures fine horizontal contours. To resolve single-view vertical ambiguity without overwriting fine XY geometry, we apply lightweight \emph{height correction} using recovered $(P_i,h_i)$---linearly rescaling occupied Z coordinates to match $h_i/\sqrt{\operatorname{Area}(P_i)}$ in the model frame before generating Stage-2/3 geometry and materials for $M_i$ (Appendix~D.5).

\paragraph{Height-guided background generation.}
We mosaic valid projected background RGB into $I_{\mathrm{BEV}}$, with $\mathcal M$ marking missing and building-reservation regions. Completion and patch generation (Figure~\ref{fig:method-overview}) follow
\begin{equation}
    \begin{aligned}
    \widetilde I_{\mathrm{BEV}}&=C_{\mathrm{bg}}(I_{\mathrm{BEV}},\mathcal M),\\
    M_j^{\mathrm{bg}}&=G_{\mathrm{bg}}\!\left(\operatorname{Crop}_j(\widetilde I_{\mathrm{BEV}});\mathbf H_j\right).
    \end{aligned}
    \label{eq:background-generation}
\end{equation}
Here $C_{\mathrm{bg}}$ applies GPT-image-2 completion and blends overlapping outputs, repairing gaps and projection artifacts while filling reservation regions with empty ground. For patch $j$, Pixal3D~\citep{li2026pixal3d} $G_{\mathrm{bg}}$ uses the completed RGB and recovered ground and elevated height fields $\mathbf H_j$, including validity and confidence, to constrain support near valid heights at sparse and fine geometry stages. Registered patches are fused in world coordinates into a city-wide background, with continuous terrain filling remaining gaps (Appendices~E.2--E.3).

\paragraph{Metric registration and CityCode assembly.}
Each building mesh is aligned to the recovered footprint, orientation, height, and base elevation. Placement checks evaluate ground contact, road intrusion, and neighboring-building clearance. The asset and transform are bound to the original entity with construction evidence, and multi-view isolated asset embeddings provide $\mathcal E_i^{(r)}$ for revision $r$.

\subsection{Dependency-Aware Versioned Execution}
\label{sec:versioned-execution}

CityCode supports tool-mediated interaction~\citep{yao2023react} over an evolving city by coupling revision-bound queries with dependency-aware, validated edits (Figure~\ref{fig:agentic-tasks}; schemas and rules in Appendix~F).

\paragraph{Revision-grounded interaction.}
Each request specifies an operation, arguments, and a read revision. Queries leave that snapshot unchanged: measurements use its entity geometry, routing its road graph, and visibility its transformed meshes. Edits must target the current revision; stale requests are rejected with replanning feedback, and outputs remain linked to their read revision.

\paragraph{Dependency-aware updates.}
An edit produces a candidate updated through operation-specific dependency rules: building edits synchronize geometry, asset transforms, block membership, and spatial relations, while road edits update the graph, blocks, affected building associations, and background layers. Checks on geometry, road clearance, and unselected entities precede publication, returning revision feedback on failure. After publication, geometric queries remain active while semantic queries wait until the updated appearance index matches the published revision.

\paragraph{Historical restoration and continued execution.}
Accepted edits publish immutable snapshots with receipts linking read and resulting revisions, with request identifiers preventing duplicate commits on retry. Historical queries inspect earlier snapshots; restoration verifies referenced resources and publishes the selected snapshot as a new revision while preserving intervening history for continued execution.

%% file: sections/04_experiments.tex
\input{sections/experiment_floats}

\section{Experiments}
\setcounter{topnumber}{3}
\label{sec:experiments}

\subsection{Experimental Setup}
\label{sec:experimental-setup}

We construct CS-Aerial from \rev{two separate maps of} the Unreal Engine City Sample project~\citep{epic2022citysample}, comprising rendered aerial RGB images, camera intrinsics and poses, and simplified 3D urban layout annotations. \rev{BigCity (sampled area about 22.58\,km$^2$) provides rendered RGB--depth pairs used only for depth adaptation and checkpoint selection; all CS-Aerial test data come from SmallCity (about 2.79\,km$^2$), which shares no area with BigCity (Appendix~C.1).} We also evaluate on the three largest UrbanScene3D scenes by area~\citep{lin2022urbanscene3d}: Qingdao Square, Qingdao Hospital, and Shenzhen University Campus, rendering observations from their reconstructed textured meshes. Reconstruction uses one aerial RGB image per approximately $400\!\times\!400$\,m \rev{region}, at $2560\!\times\!2560$ for CS-Aerial and $1024\!\times\!1024$ for UrbanScene3D. \rev{Training and inference use about eight NVIDIA A100 GPUs; reconstructing one $400\!\times\!400$\,m region averages 32\,min and \$1.00 in API calls with no manual editing (Appendix~C.1).}
Data preparation and structural extraction details are provided in Appendix~C.

Our baselines include GaussianCity~\citep{xie2025gaussiancity}, CityDreamer4D (GoogleEarth)~\citep{xie2026citydreamer4d}, Sat3DGen~\citep{qian2026sat3dgen}, Sat2Density++~\citep{qian2026sat2densitypp}, UrbanWorld2.0~\citep{wang2025urbanworld2}, and MajutsuCity~\citep{huang2025majutsucity}. Visual metrics cover pixel and structural fidelity (PSNR/SSIM), perceptual similarity (LPIPS, AlexNet), and rendered-image distributions (FID/KID). Building metrics include foreground RGB MAE, DINOv2 feature similarity, and footprint and 2.5D prism IoU after similarity alignment. Method-specific inputs, rendering, and evaluation protocols are detailed in Appendix~G.

\subsection{City and Building Generation Quality}
\label{sec:city-construction-quality}
\label{sec:building-generation-quality}

Although structure-conditioned baselines receive reference (GT) layouts/heights on CS-Aerial and shared predicted layouts on UrbanScene3D for target-scene evaluation, CityCode (aerial RGB only) ranks first on all seven CS-Aerial metrics (+1.17\,dB PSNR) and six of seven UrbanScene3D metrics (SSIM 0.286 vs.\ 0.290; Table~\ref{tab:main-results}). On 1,000/217 views ($1024\!\times\!1024$; 300/217 for VLM), CityCode leads all four VLM columns under GLM-5.3-Flash (by 2.91--50.77 points) and three additional judges (Table~S17, Appendix~G.9).

CityCode more closely matches reference building massing, street layout, and appearance (Figure~\ref{fig:scene-qualitative}), consistent with the quantitative results. Recovered structure and observed appearance jointly ground generative completion in aerial evidence, while structure-aware 3D generation helps preserve building shape and proportions. Although UrbanWorld2.0 and Sat3DGen also use input imagery, their outputs show weaker correspondence to the observed scene, suggesting less effective use of this evidence to constrain generation. Ground regions also show simplified road and vegetation patterns in UrbanWorld2.0 and blurred surface textures in Sat3DGen.

\paragraph{Building-level fidelity.}
We compare CityCode with MajutsuCity and UrbanWorld2.0 on 547 building instances. CityCode leads all appearance metrics, improving PSNR by 3.47 and 3.67\,dB and reducing foreground RGB MAE by 39.8\% and 36.1\%, respectively. These gains are consistent with our completion strategy, which combines observed source and roof appearance with recovered geometry to guide building proportions. Its 2.5D prism IoU is 0.737, slightly below MajutsuCity's 0.746 but above UrbanWorld2.0's 0.394 (Appendix~G.2).

\subsection{Metric Depth and City Geometry}
\label{sec:metric-depth-city-geometry}

Since Sat3DGen~\citep{qian2026sat3dgen} and Sat2Density++~\citep{qian2026sat2densitypp} lack instance building dimensions, Table~\ref{tab:metric-geometry} compares their mesh surfaces with CityCode's depth-derived surface before proxy construction. CityCode leads all four DSM metrics on both datasets. Relative to the strongest baseline, RMSE decreases from 33.99 to 14.37\,m on CS-Aerial and from 19.88 to 12.87\,m on UrbanScene3D, without real-scene fine-tuning. We calibrate CityCode-Depth predictions before backprojection using road and hardscape anchors, camera geometry, and local ground elevations. Without depth-scale calibration, CityCode still achieves lower DSM RMSE than both baselines on CS-Aerial (Appendix~G.4).

With other stages fixed, CityCode-Depth leads in proxy footprint IoU, XY error, and height MAE (Table~\ref{tab:building-depth-ablation}), linking depth accuracy to building recovery. Under survey-grade and field-flight camera pose and intrinsic noise, building F1 remains unchanged and geometry shifts by at most 0.31\,m (Table~S18, Appendix~G.10). Beyond buildings, our recovered road graph reaches 0.886 ground APLS (0.899 TOPO F1) and matches 89.8\% of reference shortest routes across 2,000 pairs (Table~S19, Appendix~G.11).

\subsection{Long-Horizon Urban Interaction}
\label{sec:long-horizon-interaction}

Beyond the example in Figure~\ref{fig:agentic-tasks}, we evaluate long-horizon interaction under two complementary protocols. Under a \emph{strict built-in protocol} across 50 task chains of 50 steps each---using only each method's native interfaces and stopping at the first failed or unsupported step---CityCode completes all 50/50 chains and passes all 2,500 steps, whereas neither UrbanWorld2.0 nor MajutsuCity completes a chain due to cross-object clearance violations, missing road-relation workflows, and the absence of integrated historical state restoration (Table~S14, Appendix~G.7).

To separate representation operability from missing built-in tools, we further evaluate a \emph{relaxed agent-scripted protocol} on ten 50-step chains where the same coding agent may write custom Python/Blender workflows and file-copy snapshots directly on the baselines' native scenes, retry up to three times per step, and restore step-start states after failures and continue (Table~S21, Appendix~G.13). Even under this relaxed protocol, CityCode passes all 10/10 chains with 0 constraint-violating edit attempts in 408\,min, whereas UrbanWorld2.0 and MajutsuCity pass only 3/10 chains each, incur 20 and 28 constraint-violating edit attempts, and require 3.0--3.6$\times$ the agent time (1,204 and 1,467\,min). Spatial and routing queries executed directly on our reconstructed city also match reference-city answers in 83.2--92.1\% of cases, with 1.36\,m median inter-building distance error (Table~S20, Appendix~G.12).

\subsection{Ablations and Analysis}
\label{sec:ablations-analysis}

We cross BigCity depth adaptation with scale calibration while fixing segmentation and structure recovery. Table~\ref{tab:depth-scale-ablation} reports camera-depth RMSE averaged over 79 images, distinct from the DSM elevation RMSE in Table~\ref{tab:metric-geometry}. Adaptation reduces it from 69.58 to 8.75\,m without calibration, and combining both reaches 6.27\,m. On the shared subset, combining both gives the best geometry: height MAE is 3.38\,m versus 4.17\,m (adaptation alone) and 6.98\,m (calibration alone).

Table~\ref{tab:building-ablation} ablates structure- and observation-guided building generation (Section~\ref{sec:building-generation}). Bypassing \emph{appearance completion} drops footprint and prism IoU sharply (to 0.7356 and 0.6190), while omitting proxy \emph{structure guidance} during completion lowers footprint IoU from 0.8423 to 0.7920. Replacing \emph{CityCode-Building} with \emph{Official TRELLIS.2} retains high DINOv2 similarity (0.5824) but drops 2.5D prism IoU to 0.6616: within CityCode-Building, \emph{height correction} drives vertical agreement (raising prism IoU from 0.6641 to 0.7373), and \emph{auxiliary structure losses} further improve footprint and prism IoU at a slight PSNR trade-off. Finally, replacing the two closed-source stages with open-weight SAM~3.1 and FLUX.2-klein-9B preserves scene geometry (DSM RMSE 14.39 vs.\ 14.37\,m) and building shape (prism IoU 0.775 vs.\ 0.737) at a moderate facade PSNR cost (12.65 vs.\ 14.46\,dB), confirming that geometric fidelity stems from structural coupling rather than proprietary weights (Tables~S15--S16, Appendix~G.8).

%% file: sections/experiment_floats.tex
\input{tables/main_results}

\begin{figure*}[t]
    \centering
    \includegraphics[width=\linewidth]{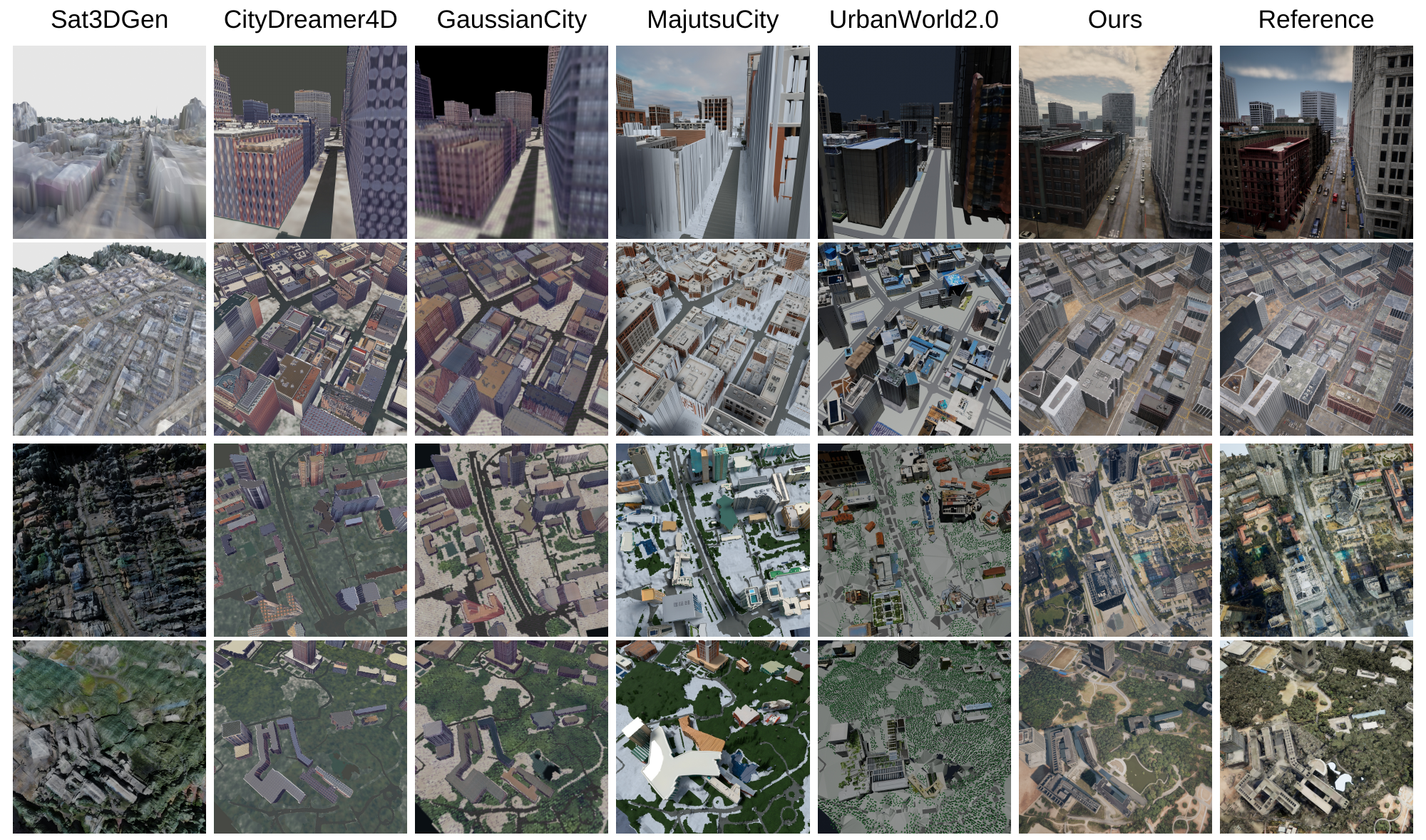}
    \caption{\textbf{City-level qualitative comparison.} Top two rows: CS-Aerial; bottom two: Qingdao Hospital and Shenzhen University Campus. Reference denotes renders of the source scenes.}
    \label{fig:scene-qualitative}
\end{figure*}

\input{tables/metric_geometry}

\input{tables/building_depth_ablation}

\begin{figure*}[t]
    \centering
    \includegraphics[width=\linewidth]{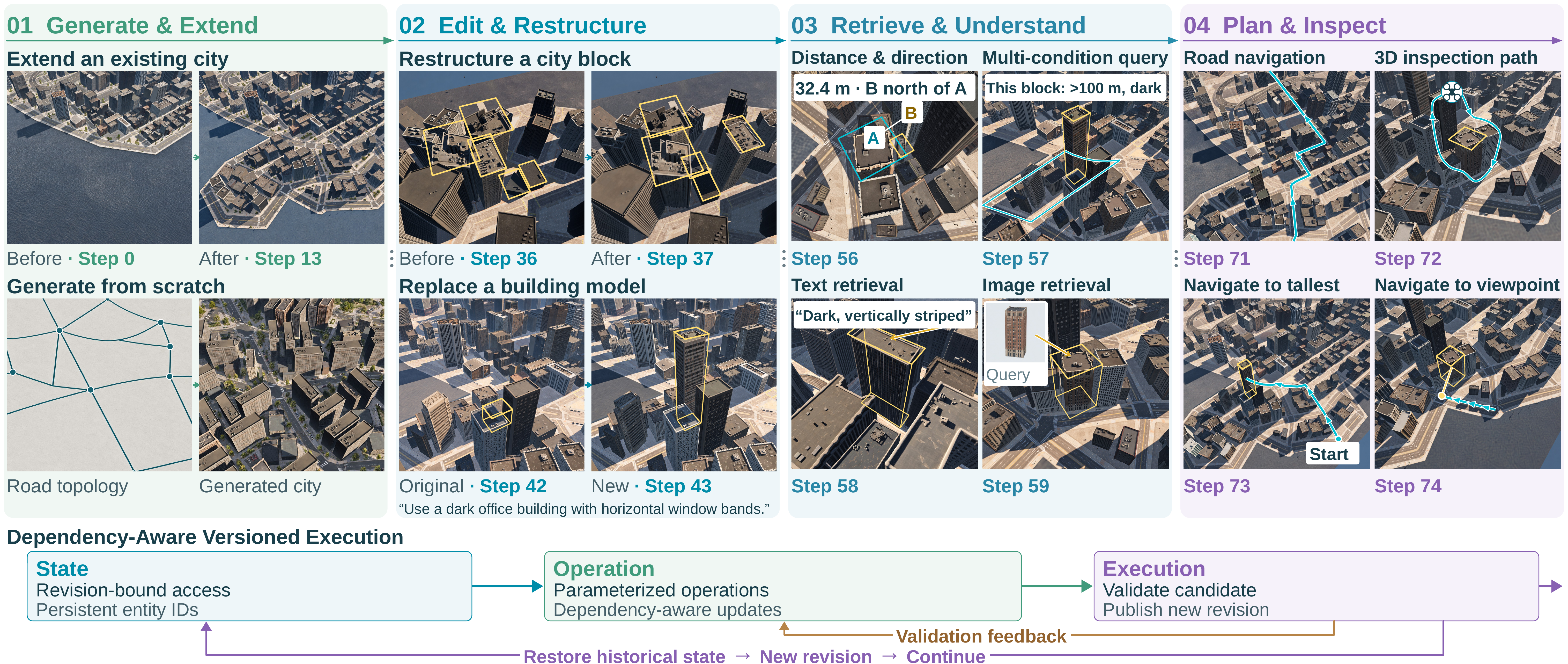}
    \caption{\textbf{Long-horizon urban interaction.} Selected checkpoints from a 116-step history show city extension, editing, retrieval, spatial queries, and planning. Step numbers identify checkpoints. The unnumbered from-scratch panel is an additional construction example. The lower strip shows versioned execution, validation feedback, and restoration followed by continued interaction.}
    \label{fig:agentic-tasks}
\end{figure*}

\input{tables/air2code_ablations}

%% file: tables/main_results.tex
\begin{table*}[t]
  \caption{\textbf{City-level visual quality on CS-Aerial and UrbanScene3D.} KID is $\times10^3$; GLM-5.3-Flash preference is in \% (Appendix~G.1). \smash{$^\dagger$}Conditioned on reference (GT) layouts/heights on CS-Aerial and shared predicted layouts on UrbanScene3D; others use only aerial RGB.}
  \label{tab:main-results}
  \centering
  \fontsize{8.3}{9.3}\selectfont
  \setlength{\tabcolsep}{1pt}
  \resizebox{\linewidth}{!}{%
  \begin{tabular}{@{}lrrrrrrr@{\hspace{4pt}}rrrrrrr@{}}
    \toprule
    & \multicolumn{7}{c}{CS-Aerial} & \multicolumn{7}{c}{UrbanScene3D} \\
    \cmidrule(lr){2-8}\cmidrule(lr){9-15}
    Method & PSNR$\uparrow$ & SSIM$\uparrow$ & LPIPS$\downarrow$ & FID$\downarrow$ & KID$\downarrow$
           & \shortstack{Visual\\pref.$\uparrow$} & \shortstack{GT-cond.\\pref.$\uparrow$}
           & PSNR$\uparrow$ & SSIM$\uparrow$ & LPIPS$\downarrow$ & FID$\downarrow$ & KID$\downarrow$
           & \shortstack{Visual\\pref.$\uparrow$} & \shortstack{GT-cond.\\pref.$\uparrow$} \\
    \midrule
    GaussianCity\smash{$^\dagger$} & 11.21 & 0.352 & 0.553 & 128.45 & 83.98 & 47.16 & 57.25 & 8.88 & 0.202 & 0.746 & 201.80 & 160.20 & 42.01 & 40.05 \\
    CityDreamer4D\smash{$^\dagger$} & 10.91 & 0.308 & 0.534 & 133.71 & 87.14 & 63.23 & 71.04 & 10.75 & \textbf{0.290} & 0.688 & 197.88 & 143.10 & 57.69 & 44.08 \\
    Sat3DGen & 11.79 & 0.333 & 0.584 & 235.74 & 190.87 & 13.38 & 12.70 & 9.24 & 0.182 & 0.625 & 172.39 & 115.40 & 13.86 & 45.39 \\
    Sat2Density++ & 11.57 & 0.352 & 0.579 & 235.71 & 196.04 & 6.04 & 7.99 & 8.48 & 0.164 & 0.626 & 167.90 & 115.70 & 5.84 & 41.78 \\
    UrbanWorld2.0\smash{$^\dagger$} & 11.58 & 0.360 & 0.553 & 148.80 & 109.23 & 56.52 & 36.21 & 9.02 & 0.184 & 0.637 & 209.64 & 163.60 & 61.48 & 47.35 \\
    MajutsuCity\smash{$^\dagger$} & 11.85 & 0.357 & 0.523 & 126.61 & 81.98 & 69.12 & 80.91 & 9.26 & 0.280 & 0.686 & 202.95 & 140.90 & 69.91 & 33.21 \\
    \midrule
    CityCode (Ours) & \textbf{13.02} & \textbf{0.453} & \textbf{0.466} & \textbf{105.21} & \textbf{67.66}
                    & \textbf{94.71} & \textbf{83.82}
                    & \textbf{11.48} & 0.286 & \textbf{0.538} & \textbf{116.73} & \textbf{47.70} & \textbf{99.23} & \textbf{98.12} \\
    \bottomrule
  \end{tabular}%
  }
\end{table*}

%% file: tables/metric_geometry.tex
\begin{table*}[t]
  \centering
  \caption{\textbf{Pre-structuring surface geometry.}
  All methods use known cameras and ground-elevation priors for metric adaptation.
  Surface elevation is evaluated on a common 1\,m DSM grid after mean fusion
  over the same regions. Input views and adaptation procedures are detailed in
  Appendix~G.4.}
  \label{tab:metric-geometry}
  \small
  \setlength{\tabcolsep}{3pt}
  \resizebox{\linewidth}{!}{%
  \begin{tabular}{lrrrr@{\hspace{12pt}}rrrr}
    \toprule
    & \multicolumn{4}{c}{CS-Aerial (synthetic)}
    & \multicolumn{4}{c}{UrbanScene3D (sampled real-world scenes)} \\
    \cmidrule(lr){2-5}\cmidrule(lr){6-9}
    Method & DSM MAE$\downarrow$ & DSM RMSE$\downarrow$ & Acc@2.5m$\uparrow$ & Acc@7.5m$\uparrow$
           & DSM MAE$\downarrow$ & DSM RMSE$\downarrow$ & Acc@2.5m$\uparrow$ & Acc@7.5m$\uparrow$ \\
    & (m) & (m) & (\%) & (\%) & (m) & (m) & (\%) & (\%) \\
    \midrule
    Sat3DGen & 23.44 & 37.03 & 15.65 & 34.99 & 9.60 & 19.88 & 32.00 & 64.91 \\
    Sat2Density++ & 17.37 & 33.99 & 15.03 & 54.36 & 10.01 & 20.97 & 30.98 & 66.29 \\
    CityCode (Ours) & \textbf{6.08} & \textbf{14.37} & \textbf{62.29} & \textbf{81.37}
                    & \textbf{7.38} & \textbf{12.87} & \textbf{33.94} & \textbf{68.04} \\
    \bottomrule
  \end{tabular}%
  }
\end{table*}

%% file: tables/building_depth_ablation.tex
\begin{table}[t]
  \centering
  \caption{\textbf{Depth backbone comparison within CityCode}, with fixed
  segmentation, calibration, and structure recovery. Evaluation uses 339 common
  CS-Aerial building pairs (Appendix~G.5).}
  \label{tab:building-depth-ablation}
  \small
  \setlength{\tabcolsep}{7pt}
  \resizebox{\linewidth}{!}{%
  \begin{tabular}{@{}lrrr@{}}
    \toprule
    Depth model & Proxy footprint & XY error & Height MAE \\
                & IoU$\uparrow$ & (m)$\downarrow$ & (m)$\downarrow$ \\
    \midrule
    MoGe-2~\citep{wang2025moge2} & 0.7584 & 3.09 & 8.06 \\
    MoGe-3~\citep{kong2026moge3} & 0.7858 & 2.69 & 6.80 \\
    MoGe2-Aerial & 0.7700 & 2.82 & 6.12 \\
    CityCode-Depth & \textbf{0.8446} & \textbf{1.58} & \textbf{2.92} \\
    \bottomrule
  \end{tabular}}
\end{table}

%% file: tables/air2code_ablations.tex
\begin{table*}[t]
  \centering
  \caption{\textbf{Ablations of Air2Code components.}
  (a) Adapted or calibrated configurations share 445 matched pairs.
  (b) 547 buildings, 48 views each; GT footprints and heights define reference prisms.
  Protocols: Appendices~G.6
  and~G.3.}
  \label{tab:air2code-ablations}
  \normalsize
  \input{tables/depth_scale_ablation}
  \par\vspace{7pt}
  \input{tables/building_ablation}
\end{table*}

%% file: tables/depth_scale_ablation.tex
\begingroup
  \textbf{(a) Metric structure recovery}\par\vspace{3pt}
  \makeatletter
  \edef\@currentlabel{\thetable(a)}\label{tab:depth-scale-ablation}
  \makeatother
  \setlength{\tabcolsep}{4pt}
  \begin{tabular*}{\linewidth}{@{\extracolsep{\fill}}ccrrrr@{}}
    \toprule
    Depth & Scale & Camera-depth & Proxy footprint & XY error & Height MAE \\
    adaptation & calibration & RMSE (m)$\downarrow$ & IoU$\uparrow$ & (m)$\downarrow$ & (m)$\downarrow$ \\
    \midrule
    $\times$ & $\times$ & 69.58 & 0.4831 & 5.85 & 62.55 \\
    $\checkmark$ & $\times$ & 8.75 & 0.7723 & 2.73 & 4.17 \\
    $\times$ & $\checkmark$ & 11.84 & 0.7153 & 3.65 & 6.98 \\
    $\checkmark$ & $\checkmark$ & \textbf{6.27} & \textbf{0.8039} & \textbf{2.18} & \textbf{3.38} \\
    \bottomrule
  \end{tabular*}
\endgroup

%% file: tables/building_ablation.tex
\begingroup
  \textbf{(b) Generated building quality}\par\vspace{3pt}
  \makeatletter
  \edef\@currentlabel{\thetable(b)}\label{tab:building-ablation}
  \makeatother
  \setlength{\tabcolsep}{2.5pt}
  \begin{tabular*}{\linewidth}{@{\extracolsep{\fill}}lrrrrrr@{}}
    \toprule
    Configuration & Footprint & 2.5D prism & PSNR & SSIM & DINOv2 & RGB \\
     & IoU$\uparrow$ & IoU$\uparrow$ & (dB)$\uparrow$ & $\uparrow$ & $\uparrow$ & MAE$\downarrow$ \\
    \midrule
    Full & \textbf{0.8423} & \textbf{0.7373} & 14.4586 & \textbf{0.7706} & 0.5678 & 0.1611 \\
    w/o appearance completion & 0.7356 & 0.6190 & 13.4382 & 0.7152 & 0.4312 & 0.1767 \\
    Completion w/o structure guidance & 0.7920 & 0.6710 & 14.0885 & 0.7633 & 0.5605 & 0.1624 \\
    Official TRELLIS.2 & 0.8404 & 0.6616 & 13.6701 & 0.7639 & \textbf{0.5824} & 0.1743 \\
    w/o height correction & 0.8413 & 0.6641 & 14.4206 & 0.7687 & 0.5596 & \textbf{0.1602} \\
    w/o auxiliary structure losses & 0.8341 & 0.7295 & \textbf{14.5252} & 0.7697 & 0.5617 & 0.1621 \\
    \bottomrule
  \end{tabular*}
\endgroup

%% file: sections/05_conclusion.tex
\section{Conclusion}
\label{sec:conclusion}

CityCode and Air2Code build grounded urban worlds from sparse aerial observations. Recovered metric building instances and road structure link observations, generated assets, and spatial operations through persistent identities, while versioned execution keeps the city state consistent across edits and restoration. Across CS-Aerial and UrbanScene3D, CityCode leads on 21 of 22 metric-geometry and visual-quality metrics while preserving geometric fidelity with open-weight backbones. Across 50 chains of 50 steps, it completes every chain and outperforms agent-scripted baselines with zero constraint violations at 3.0--3.6$\times$ lower agent time.

\textbf{Limitations and future work.} First, peak facade appearance still relies on closed-source image models: open-weight replacements match or improve building geometry but reproduce observed facades less faithfully. Second, a single view per region can miss occluded structure such as roads under elevated decks. Future work will explore finer geometry, larger cities, and applications to what-if urban planning and embodied aerial navigation or inspection.

%% file: sections/llm_usage.tex
\section*{LLM Usage Statement}
We used OpenAI Codex for writing assistance, literature review, code debugging, result checking, and LaTeX formatting; GPT-5.6-sol and GPT-6.1-sol for long-horizon evaluation; and GLM-5.3-Flash, DeepSeek V4.1 Flash, Qwen3-VL-32B, and Gemini~4 Argon for visual judging. Our method uses Gemini~3 Pro Image for mask extraction, and GPT-image-2 and FLUX.2-klein-9B for appearance completion. Method and evaluation details are provided in the main text and supplementary material.

%% file: main.bbl
\begin{thebibliography}{47}
\providecommand{\natexlab}[1]{#1}
\providecommand{\url}[1]{\texttt{#1}}
\expandafter\ifx\csname urlstyle\endcsname\relax
  \providecommand{\doi}[1]{doi: #1}\else
  \providecommand{\doi}{doi: \begingroup \urlstyle{rm}\Url}\fi

\bibitem[Bernard et~al.(2022)Bernard, Bocher, {Le Saux Wiederhold}, Leconte,
  and Masson]{bernard2022buildingheight}
J{\'e}r{\'e}my Bernard, Erwan Bocher, Elisabeth {Le Saux Wiederhold}, Fran{\c
  c}ois Leconte, and Val{\'e}ry Masson.
\newblock Estimation of missing building height in {OpenStreetMap} data: a
  {French} case study using {GeoClimate} 0.0.1.
\newblock \emph{Geoscientific Model Development}, 15:\penalty0 7505--7532,
  2022.

\bibitem[Caldarelli et~al.(2023)Caldarelli, Arcaute, Barthelemy, Batty,
  Gershenson, Helbing, Mancuso, Moreno, Ramasco, Rozenblat, S{\'a}nchez, and
  Fern{\'a}ndez-Villaca{\~n}as]{caldarelli2023complexity}
G. Caldarelli, E. Arcaute, M. Barthelemy, M. Batty, C. Gershenson, D. Helbing,
  S. Mancuso, Y. Moreno, J.~J. Ramasco, C. Rozenblat, A. S{\'a}nchez, and J.~L.
  Fern{\'a}ndez-Villaca{\~n}as.
\newblock The role of complexity for digital twins of cities.
\newblock \emph{Nature Computational Science}, 3:\penalty0 374--381, 2023.

\bibitem[Chen et~al.(2026)Chen, Lin, and Zhang]{chen2026codeworldmodel}
Yiwen Chen, Guosheng Lin, and Chi Zhang.
\newblock {Code World Model: Coding Agent as World Brain}.
\newblock \emph{arXiv preprint arXiv:2608.25927}, 2026.

\bibitem[Chung et~al.(2026)Chung, Lee, Xiang, Yang, and
  Lee]{chung2026map2world}
Jaeyoung Chung, Suyoung Lee, Jianfeng Xiang, Jiaolong Yang, and Kyoung~Mu Lee.
\newblock {Map2World: Segment Map Conditioned Text to 3D World Generation}.
\newblock \emph{arXiv preprint arXiv:2605.00781}, 2026.

\bibitem[Deng et~al.(2024)Deng, Chai, Huang, Zhao, Huang, Gao, Guo, Hao, Hu,
  Hwang, Li, and Wang]{deng2024citycraft}
Jie Deng, Wenhao Chai, Junsheng Huang, Zhonghan Zhao, Qixuan Huang, Mingyan
  Gao, Jianshu Guo, Shengyu Hao, Wenhao Hu, Jenq-Neng Hwang, Xi Li, and Gaoang
  Wang.
\newblock {CityCraft: A Real Crafter for 3D City Generation}.
\newblock \emph{arXiv preprint arXiv:2406.04983}, 2024.

\bibitem[{Epic Games}(2022)]{epic2022citysample}
{Epic Games}.
\newblock {City Sample}.
\newblock Unreal Engine sample project, 2022.

\bibitem[Feng et~al.(2025)Feng, Zhang, Liu, Zhang, Ouyang, Yan, Du, Guo, and
  Li]{feng2025citybench}
Jie Feng, Jun Zhang, Tianhui Liu, Xin Zhang, Tianjian Ouyang, Junbo Yan, Yuwei
  Du, Siqi Guo, and Yong Li.
\newblock {CityBench: Evaluating the Capabilities of Large Language Models for
  Urban Tasks}.
\newblock In \emph{Proceedings of the ACM SIGKDD Conference on Knowledge
  Discovery and Data Mining}, 2025.

\bibitem[Gu et~al.(2023)Gu, Kuwajerwala, Morin, Jatavallabhula, Sen, Agarwal,
  Rivera, Paul, Ellis, Chellappa, Gan, de~Melo, Tenenbaum, Torralba, Shkurti,
  and Paull]{gu2023conceptgraphs}
Qiao Gu, Alihusein Kuwajerwala, Sacha Morin, Krishna~Murthy Jatavallabhula,
  Bipasha Sen, Aditya Agarwal, Corban Rivera, William Paul, Kirsty Ellis, Rama
  Chellappa, Chuang Gan, Celso~Miguel de Melo, Joshua~B. Tenenbaum, Antonio
  Torralba, Florian Shkurti, and Liam Paull.
\newblock {ConceptGraphs}: Open-vocabulary {3D} scene graphs for perception and
  planning.
\newblock \emph{arXiv preprint arXiv:2309.16650}, 2023.

\bibitem[Herfort et~al.(2023)Herfort, Lautenbach, {Porto de Albuquerque},
  Anderson, and Zipf]{herfort2023osmcompleteness}
Benjamin Herfort, Sven Lautenbach, Jo{\~a}o {Porto de Albuquerque}, Jennings
  Anderson, and Alexander Zipf.
\newblock A spatio-temporal analysis investigating completeness and
  inequalities of global urban building data in {OpenStreetMap}.
\newblock \emph{Nature Communications}, 14:\penalty0 3985, 2023.

\bibitem[Hu et~al.(2026)Hu, Li, Yu, Liu, Ye, Chen, and Huang]{hu2026dgbrf}
Zhenghao Hu, Weijia Li, Jinhua Yu, Minfa Liu, Junyan Ye, Peimin Chen, and
  Huabing Huang.
\newblock {3D} building reconstruction from monocular remote sensing imagery
  via diffusion models and geometric priors.
\newblock \emph{ISPRS Journal of Photogrammetry and Remote Sensing},
  232:\penalty0 124--137, 2026.

\bibitem[Hua et~al.(2025)Hua, Jiang, Chen, and Zhao]{hua2025sat2city}
Tongyan Hua, Lutao Jiang, Ying-Cong Chen, and Wufan Zhao.
\newblock {Sat2City}: {3D} city generation from a single satellite image with
  cascaded latent diffusion.
\newblock \emph{arXiv preprint arXiv:2507.04403}, 2025.

\bibitem[Hua et~al.(2026)Hua, Wu, Zhu, Ren, Hong, Chen, Xiong, and
  Zhao]{hua2026sat2cityv2}
Tongyan Hua, Dongli Wu, Jinjing Zhu, Yinrui Ren, Zhongcheng Hong, Ying-Cong
  Chen, Hui Xiong, and Wufan Zhao.
\newblock {Sat2City v2: Native 3D City Asset Generation from a Single Satellite
  Image}.
\newblock \emph{arXiv preprint arXiv:2606.24138}, 2026.

\bibitem[Huang et~al.(2024)Huang, Yong, Ma, Linghu, Li, Wang, Li, Zhu, Jia, and
  Huang]{huang2024leo}
Jiangyong Huang, Silong Yong, Xiaojian Ma, Xiongkun Linghu, Puhao Li, Yan Wang,
  Qing Li, Song-Chun Zhu, Baoxiong Jia, and Siyuan Huang.
\newblock {An Embodied Generalist Agent in 3D World}.
\newblock In \emph{International Conference on Machine Learning}, 2024.

\bibitem[Huang et~al.(2025)Huang, He, Huang, Xiong, Luo, Ye, Li, Chen, and
  Han]{huang2025majutsucity}
Zilong Huang, Jun He, Xiaobin Huang, Ziyi Xiong, Yang Luo, Junyan Ye, Weijia
  Li, Yiping Chen, and Ting Han.
\newblock {MajutsuCity}: Language-driven aesthetic-adaptive city generation
  with controllable {3D} assets and layouts.
\newblock \emph{arXiv preprint arXiv:2511.20415}, 2025.

\bibitem[Hughes et~al.(2022)Hughes, Chang, and Carlone]{hughes2022hydra}
Nathan Hughes, Yun Chang, and Luca Carlone.
\newblock {Hydra: A Real-time Spatial Perception System for 3D Scene Graph
  Construction and Optimization}.
\newblock In \emph{Robotics: Science and Systems}, 2022.

\bibitem[Jin and Feng(2025)]{jin2025satskylines}
Zhangyu Jin and Andrew Feng.
\newblock {SAT-SKYLINES}: {3D} building generation from satellite imagery and
  coarse geometric priors.
\newblock \emph{arXiv preprint arXiv:2508.18531}, 2025.

\bibitem[Juh{\'a}sz(2021)]{juhasz2021temporalaccuracy}
Levente Juh{\'a}sz.
\newblock Towards understanding the temporal accuracy of {OpenStreetMap}: A
  quantitative experiment.
\newblock In \emph{Proceedings of the Academic Track at the State of the Map
  2021}, pages 19--22, 2021.
\newblock Extended abstract.

\bibitem[Kong et~al.(2026)Kong, Li, Wang, Xu, Yao, Xiang, and
  Yang]{kong2026moge3}
Lingyu Kong, Ruicheng Li, Ruicheng Wang, Sicheng Xu, Chengtang Yao, Jianfeng
  Xiang, and Jiaolong Yang.
\newblock {MoGe-3}: Fine-detail monocular geometry estimation with self-guided
  sparse volumetric refinement.
\newblock \emph{arXiv preprint arXiv:2607.17967}, 2026.

\bibitem[Li et~al.(2026{\natexlab{a}})Li, Zhao, Chen, Hu, Guo, Zhang, Shan, and
  Hu]{li2026pixal3d}
Dong-Yang Li, Wang Zhao, Yuxin Chen, Wenbo Hu, Meng-Hao Guo, Fang-Lue Zhang,
  Ying Shan, and Shi-Min Hu.
\newblock {Pixal3D}: Pixel-aligned {3D} generation from images.
\newblock \emph{arXiv preprint arXiv:2605.10922}, 2026{\natexlab{a}}.

\bibitem[Li et~al.(2024)Li, Yang, Hu, Zheng, Xia, and He]{li2024mlsbrn}
Weijia Li, Haote Yang, Zhenghao Hu, Juepeng Zheng, Gui-Song Xia, and Conghui
  He.
\newblock {3D} building reconstruction from monocular remote sensing images
  with multi-level supervisions.
\newblock In \emph{Proceedings of the IEEE/CVF Conference on Computer Vision
  and Pattern Recognition (CVPR)}, pages 27728--27737, 2024.

\bibitem[Li et~al.(2026{\natexlab{b}})Li, Liao, and Zhu]{li2026recursive}
Zhiqi Li, Yuxuan Liao, and Bo Zhu.
\newblock Recursive code world models: Building complex worlds through
  recursive scene programs.
\newblock \emph{arXiv preprint arXiv:2609.11499}, 2026{\natexlab{b}}.

\bibitem[Lin et~al.(2022)Lin, Liu, Hu, Yan, Xie, and
  Huang]{lin2022urbanscene3d}
Liqiang Lin, Yilin Liu, Yue Hu, Xingguang Yan, Ke Xie, and Hui Huang.
\newblock {Capturing, Reconstructing, and Simulating: the UrbanScene3D
  Dataset}.
\newblock In \emph{European Conference on Computer Vision}, 2022.

\bibitem[Liu et~al.(2026)Liu, Tang, Wu, Zheng, Hu, Hui, Xie, Dai, and
  Liu]{liu2026citygenagent}
Zishan Liu, Zecong Tang, RuoCheng Wu, Xinzhe Zheng, Jingyu Hu, Ka-Hei Hui,
  Haoran Xie, Bo Dai, and Zhengzhe Liu.
\newblock {Imagine a City: CityGenAgent for Procedural 3D City Generation}.
\newblock \emph{arXiv preprint arXiv:2602.05362}, 2026.

\bibitem[Mao et~al.(2025)Mao, Zhong, Fang, Zheng, Tang, Zhu, Tan, and
  Zhou]{mao2025spatiallm}
Yongsen Mao, Junhao Zhong, Chuan Fang, Jia Zheng, Rui Tang, Hao Zhu, Ping Tan,
  and Zihan Zhou.
\newblock {SpatialLM: Training Large Language Models for Structured Indoor
  Modeling}.
\newblock \emph{arXiv preprint arXiv:2506.07491}, 2025.

\bibitem[{Open Geospatial Consortium}(2021)]{ogc2021citygml}
{Open Geospatial Consortium}.
\newblock {OGC City Geography Markup Language (CityGML) Part 1: Conceptual
  Model Standard}.
\newblock Technical Report 20-010, Open Geospatial Consortium, 2021.
\newblock Version 3.0.

\bibitem[Qian et~al.(2026{\natexlab{a}})Qian, Tan, Wang, Zheng, Xiong, Xia,
  Shen, and Xue]{qian2026sat2densitypp}
Ming Qian, Bin Tan, Qiuyu Wang, Xianwei Zheng, Hanjiang Xiong, Gui-Song Xia,
  Yujun Shen, and Nan Xue.
\newblock Seeing through satellite images at street views.
\newblock \emph{arXiv preprint arXiv:2505.17001v2}, 2026{\natexlab{a}}.
\newblock Accepted to IEEE Transactions on Pattern Analysis and Machine
  Intelligence.

\bibitem[Qian et~al.(2026{\natexlab{b}})Qian, Xia, Liu, Ma, Wang, Ke, Tan,
  Zhang, and Xia]{qian2026sat3dgen}
Ming Qian, Zimin Xia, Changkun Liu, Shuailei Ma, Wen Wang, Zeran Ke, Bin Tan,
  Hang Zhang, and Gui-Song Xia.
\newblock {Sat3DGen}: Comprehensive street-level {3D} scene generation from
  single satellite image.
\newblock \emph{arXiv preprint arXiv:2605.14984}, 2026{\natexlab{b}}.

\bibitem[Rana et~al.(2023)Rana, Haviland, Garg, Abou-Chakra, Reid, and
  Suenderhauf]{rana2023sayplan}
Krishan Rana, Jesse Haviland, Sourav Garg, Jad Abou-Chakra, Ian Reid, and Niko
  Suenderhauf.
\newblock {SayPlan: Grounding Large Language Models using 3D Scene Graphs for
  Scalable Robot Task Planning}.
\newblock In \emph{Conference on Robot Learning}, 2023.

\bibitem[Remondino et~al.(2011)Remondino, Barazzetti, Nex, Scaioni, and
  Sarazzi]{remondino2011uav}
F. Remondino, L. Barazzetti, F. Nex, M. Scaioni, and D. Sarazzi.
\newblock {UAV} photogrammetry for mapping and {3D} modeling -- current status
  and future perspectives.
\newblock \emph{The International Archives of the Photogrammetry, Remote
  Sensing and Spatial Information Sciences}, XXXVIII-1/C22:\penalty0 25--31,
  2011.

\bibitem[Shang et~al.(2024)Shang, Lin, Zheng, Fan, Ding, Feng, Chen, Tian, and
  Li]{shang2024urbanworld}
Yu Shang, Yuming Lin, Yu Zheng, Hangyu Fan, Jingtao Ding, Jie Feng, Jiansheng
  Chen, Li Tian, and Yong Li.
\newblock {UrbanWorld}: An urban world model for {3D} city generation.
\newblock \emph{arXiv preprint arXiv:2407.11965}, 2024.

\bibitem[Shao et~al.(2024)Shao, Wu, Li, Peng, Chen, and
  Du]{shao2024singlerecon}
Ruizhe Shao, Jiangjiang Wu, Jun Li, Shuang Peng, Hao Chen, and Chun Du.
\newblock {SingleRecon}: Reconstructing building {3-D} models of {LoD1} from a
  single off-nadir remote sensing image.
\newblock \emph{IEEE Journal of Selected Topics in Applied Earth Observations
  and Remote Sensing}, 17:\penalty0 19588--19600, 2024.

\bibitem[Song et~al.(2026)Song, Chen, Zhang, Zou, Fu, Yuan, Huang, Cui, and
  Cao]{song2026aerialmetric}
Zhongqiang Song, Guanying Chen, Yuqi Zhang, Yin Zou, Chuanyu Fu, Zhiyuan Yuan,
  Chuan Huang, Shuguang Cui, and Xiaochun Cao.
\newblock {AerialMetric}: Benchmarking and adapting {UAV} monocular metric
  depth estimation in the real world.
\newblock \emph{arXiv preprint arXiv:2606.29716}, 2026.

\bibitem[Wang et~al.(2026)Wang, Wang, Li, Yang, Lin, Li, and
  Cheng]{wang2026scenecode}
Puyi Wang, Yuhao Wang, Linjie Li, Zhengyuan Yang, Kevin~Qinghong Lin, Yangguang
  Li, and Yu Cheng.
\newblock {SceneCode}: Executable world programs for editable indoor scenes
  with articulated objects.
\newblock \emph{arXiv preprint arXiv:2605.19587}, 2026.

\bibitem[Wang et~al.(2025{\natexlab{a}})Wang, Xu, Dong, Deng, Xiang, Lv, Sun,
  Tong, and Yang]{wang2025moge2}
Ruicheng Wang, Sicheng Xu, Yue Dong, Yu Deng, Jianfeng Xiang, Zelong Lv,
  Guangzhong Sun, Xin Tong, and Jiaolong Yang.
\newblock {MoGe-2}: Accurate monocular geometry with metric scale and sharp
  details.
\newblock In \emph{Advances in Neural Information Processing Systems},
  2025{\natexlab{a}}.

\bibitem[Wang et~al.(2025{\natexlab{b}})Wang, Zheng, Shang, He, Yu, Hangyu,
  Feng, Liao, and Li]{wang2025urbanworld2}
Shengyuan Wang, Zhiheng Zheng, Yu Shang, Lixuan He, Yangcheng Yu, Fan Hangyu,
  Jie Feng, Qingmin Liao, and Yong Li.
\newblock {UrbanWorld2.0: A Multimodal Agentic Framework for Reality-Aligned 3D
  World Generation at City-Scale}.
\newblock \emph{arXiv preprint arXiv:2511.18005}, 2025{\natexlab{b}}.
\newblock Version 3, revised August 2026.

\bibitem[Wang et~al.(2025{\natexlab{c}})Wang, Kang, Wu, and
  Shi]{wang2025sat2realcity}
Xinliang Wang, Yijie Kang, Zhenyu Wu, and Yifeng Shi.
\newblock {Sat2RealCity: Geometry-Aware and Appearance-Controllable 3D Urban
  Generation from Satellite Imagery}.
\newblock \emph{arXiv preprint arXiv:2511.11470}, 2025{\natexlab{c}}.

\bibitem[Wu et~al.(2025)Wu, He, He, Wang, Duan, Liu, Li, and
  Zhou]{wu2025metaurban}
Wayne Wu, Honglin He, Jack He, Yiran Wang, Chenda Duan, Zhizheng Liu, Quanyi
  Li, and Bolei Zhou.
\newblock {MetaUrban}: An embodied {AI} simulation platform for urban
  micromobility.
\newblock In \emph{International Conference on Learning Representations}, 2025.

\bibitem[Xiang et~al.(2025)Xiang, Chen, Xu, Wang, Lv, Deng, Zhu, Dong, Zhao,
  Yuan, and Yang]{xiang2025trellis2}
Jianfeng Xiang, Xiaoxue Chen, Sicheng Xu, Ruicheng Wang, Zelong Lv, Yu Deng,
  Hongyuan Zhu, Yue Dong, Hao Zhao, Nicholas~Jing Yuan, and Jiaolong Yang.
\newblock Native and compact structured latents for {3D} generation.
\newblock \emph{Tech report}, 2025.

\bibitem[Xie et~al.(2024)Xie, Chen, Hong, and Liu]{xie2024citydreamer}
Haozhe Xie, Zhaoxi Chen, Fangzhou Hong, and Ziwei Liu.
\newblock {CityDreamer}: Compositional generative model of unbounded {3D}
  cities.
\newblock In \emph{Proceedings of the IEEE/CVF Conference on Computer Vision
  and Pattern Recognition}, 2024.

\bibitem[Xie et~al.(2025)Xie, Chen, Hong, and Liu]{xie2025gaussiancity}
Haozhe Xie, Zhaoxi Chen, Fangzhou Hong, and Ziwei Liu.
\newblock Generative {Gaussian} splatting for unbounded {3D} city generation.
\newblock In \emph{Proceedings of the IEEE/CVF Conference on Computer Vision
  and Pattern Recognition}, pages 6111--6120, 2025.

\bibitem[Xie et~al.(2026)Xie, Chen, Hong, and Liu]{xie2026citydreamer4d}
Haozhe Xie, Zhaoxi Chen, Fangzhou Hong, and Ziwei Liu.
\newblock {Compositional Generative Model of Unbounded 4D Cities}.
\newblock \emph{IEEE Transactions on Pattern Analysis and Machine
  Intelligence}, 48\penalty0 (1):\penalty0 312--328, 2026.

\bibitem[Xu et~al.(2026)Xu, Hu, Zhu, Gao, Wang, Rao, Lu, Li, Yin, and
  Li]{xu2026citycube}
Haotian Xu, Yue Hu, Zhengqiu Zhu, Chen Gao, Ziyou Wang, Junreng Rao, Wenhao Lu,
  Weishi Li, Quanjun Yin, and Yong Li.
\newblock {CityCube: Benchmarking Cross-view Spatial Reasoning on
  Vision-Language Models in Urban Environments}.
\newblock \emph{arXiv preprint arXiv:2601.14339}, 2026.

\bibitem[Yao et~al.(2023)Yao, Zhao, Yu, Du, Shafran, Narasimhan, and
  Cao]{yao2023react}
Shunyu Yao, Jeffrey Zhao, Dian Yu, Nan Du, Izhak Shafran, Karthik Narasimhan,
  and Yuan Cao.
\newblock {ReAct: Synergizing Reasoning and Acting in Language Models}.
\newblock In \emph{International Conference on Learning Representations}, 2023.

\bibitem[Yao et~al.(2025)Yao, Wang, Wu, Ping, Zhang, and
  Xiong]{yao2025magiccity}
Xingbo Yao, Xuanmin Wang, Hao Wu, Chengliang Ping, Doudou Zhang, and Hui Xiong.
\newblock {MagicCity: Geometry-Aware 3D City Generation from Satellite Imagery
  with Multi-View Consistency}.
\newblock In \emph{Proceedings of the IEEE/CVF International Conference on
  Computer Vision}, pages 25325--25334, 2025.

\bibitem[Zhang et~al.(2024)Zhang, Zhou, Wang, Luo, Wang, Li, Zhang, and
  Peng]{zhang2024cityx}
Shougao Zhang, Mengqi Zhou, Yuxi Wang, Chuanchen Luo, Rongyu Wang, Yiwei Li,
  Zhaoxiang Zhang, and Junran Peng.
\newblock {CityX}: Controllable procedural content generation for unbounded
  {3D} cities.
\newblock \emph{arXiv preprint arXiv:2407.17572}, 2024.

\bibitem[Zhang et~al.(2026)Zhang, Xu, Wang, Moro, Yue, Wang, Chen, Liu, Fang,
  Gong, Bettencourt, Batty, and Li]{zhang2026generativeurban}
Yunke Zhang, Fengli Xu, Qi~R. Wang, Esteban Moro, Yang Yue, Huandong Wang, Bin
  Chen, Yu Liu, Dongping Fang, Peng Gong, Lu{\'i}s M.~A. Bettencourt, Michael
  Batty, and Yong Li.
\newblock {Generative AI in urban science and practice}.
\newblock \emph{Nature Cities}, 3:\penalty0 688--699, 2026.

\bibitem[Zheng et~al.(2025)Zheng, Zha, Xu, Gu, Yang, and
  Wang]{zheng2025scenefuse3d}
Kaizhi Zheng, Ruijian Zha, Zishuo Xu, Jing Gu, Jie Yang, and Xin~Eric Wang.
\newblock Constructing a {3D} scene from a single image.
\newblock \emph{arXiv preprint arXiv:2505.15765}, 2025.

\end{thebibliography}
